\documentclass[10pt,twocolumn,letterpaper]{article}

\usepackage{iccv}
\usepackage{times}
\usepackage{epsfig}
\usepackage{graphicx}
\usepackage{amsmath}
\usepackage{amssymb}
\usepackage{color}
\usepackage[table]{xcolor} 
\usepackage{ifthen}
\usepackage{multirow}
\usepackage{amsmath}
\usepackage{amssymb}
\usepackage{subcaption}
\usepackage[table]{xcolor} 

\usepackage[breaklinks=true,bookmarks=false,colorlinks = true]{hyperref}

\iccvfinalcopy 

\makeatletter
\newcommand*\titleheader[1]{\gdef\@titleheader{#1}}
\AtBeginDocument{
	\let\st@red@title\@title
	\def\@title{%
		\bgroup\normalfont\large\centering\@titleheader\par\egroup
		\vskip1.5em\st@red@title}
}
\makeatother

\title{Disguised Face Identification (DFI) with Facial KeyPoints using \\ Spatial Fusion Convolutional Network}
\titleheader{To Appear in the IEEE International Conference on Computer Vision Workshops (ICCVW) 2017}

\begin{document}


\author{Amarjot Singh\\ 
Department of Engineering \\
University of Cambridge, U.K.\\
{\tt\small as2436@cam.ac.uk}
\and
Devendra Patil\\
National Institute of Technology \\
Warangal, India \\
{\tt\small pdevendra@student.nitw.ac.in}
\and
G Meghana Reddy\\
National Institute of Technology \\
Warangal, India \\
{\tt\small rmeghana@student.nitw.ac.in}
\and
SN Omkar\\
Indian Institute of Science \\
Bangalore, India \\
{\tt\small omkar@aero.iisc.ernet.in}}

\maketitle


\begin{abstract}
Disguised face identification (DFI) is an extremely challenging problem due to the numerous variations that can be introduced using different disguises. This paper introduces a deep learning framework to first detect 14 facial key-points which are then utilized to perform disguised face identification. Since the training of deep learning architectures relies on large annotated datasets, two annotated facial key-points datasets are introduced. The effectiveness of the facial keypoint detection framework is presented for each keypoint. The superiority of the key-point detection framework is also demonstrated by a comparison with other deep networks. The effectiveness of classification performance is also demonstrated by comparison with the state-of-the-art face disguise classification methods.  
\end{abstract}

\section{Introduction}
Face identification is an important and challenging problem~\cite{faceamar,prob}. Face alterations can dramatically disguise one’s identity by including a wide variety of altered physical attributes such as wearing a wig, changing hairstyle or hair color, wearing eyeglasses, removing or growing a beard, etc~\cite{tejas}. Righi et al.~\cite{righi} concluded that the face recognition performance degraded with the effect of intentional face alterations such as the attire and hairstyles; by adding wigs and eyeglasses. 

In order to identify the face, there is an ardent need to analysis the shape of the face using facial keypoints. Only a few attempts have been made in the past to solve this task. Tejas et al~\cite{tejas} proposed localized feature descriptors to identify disguised face patches and used this information to improve face identification performance. Singh et al. \cite{singh2008recognizing} used texture based features to classify disguised faces.

The use of facial key-point for applications like facial expression classification, facial alignment, tracking faces in videos etc has recently gained popularity~\cite{stanford}. Numerous attempts have been made in the past to achieve this goal which narrows down to two main state-of-the-art methods. The first kind use feature extraction algorithms like Gabor features with texture-based and shape-based features to detect different facial key-points~\cite{vuka}. The second class of methods make use of probabilistic graphical models~\cite{martinez2013local} to capture the relationship between pixels and features to detect facial key-points. 

The superior performance of deep networks in different computer vision tasks have motivated the use of deep networks for facial key-point detection~\cite{sun2013deep,haavisto2013deep}. Sun et al. \cite{sun2013deep} defined a three-layer architecture which captures the global high-level features, and then refine the initialization to locate the positions of key-points. Pre-trained Deep Belief Networks (DBN) on surrounding feed-forward neural network with linear Gaussian output layer was used by Haavisto et al. \cite{haavisto2013deep} to detect facial key points. 

The use of deep network for this application is challenging as the amount of annotated training data required to train the deep networks is not available (small: AR~\cite{ar} and Yale~\cite{yale} face databases) for this application forcing the designers to use transfer learning. Transfer learning often performs well but may under-perform as the amount of training data is may not sufficient to fine-tune the pre-trained deep networks. 

This paper introduces a facial key-point detection framework for disguised face identification. The framework first uses a deep convolutional network to detect 14 facial key-points, as shown in Fig. 1, that were identified as essential for facial identification~\cite{stanford}. The detected points are then connected to form a star-net structure (Fig. 3). The orientations between the connected points in the star-net structure are then used by a proposed classification framework to perform the facial identification. The paper also introduces two annotated facial disguise datasets to improve the training of deep convolutional network due to their reliance on large training datasets. 

\begin{figure}[t!] 
\centering    
\includegraphics[scale = 0.32]{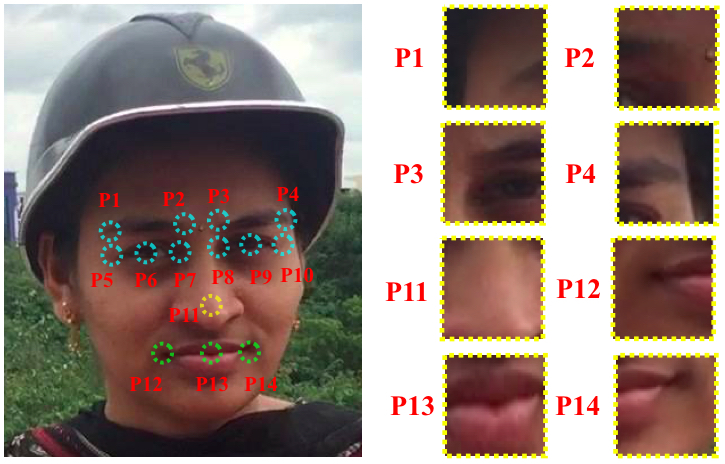}
\caption{{The figure (left) illustrates the 14 facial key-points annotated for both the introduced datasets. The description of the facial points is as: Eyes region (cyan): P1- left eyebrow outer corner, P2- left eyebrow inner corner, P3- right eyebrow inner corner, P4- right eyebrow outer corner, P5- left eye outer corner, P6- left eye center, P7- left eye inner corner, P8- right eye inner corner, P9- right eye center, P10- right eye outer corner; Nose region (yellow): P11- nose; Lip region (green) P12- lip left corner, P13- lip centre, P14- lip right corner. Few key points have been shown on the right.  }}
\label{fig:scatter00}
\end{figure}

The main contributions of the paper are stated below:

\begin{itemize}

\item{\textit{Disguised Face Identification (DFI) Framework}:} The proposed framework uses a Spatial Fusion deep convolutional network~\cite{fusion} to extract 14 key-points from the face that are considered essential to describe the facial structure~\cite{stanford}. The extracted points are connected to form a star-net structure (Fig. 3). The orientations of the connected points are used by the proposed classification framework (Section 4) for face identification. 

\item{\textit{Simple and Complex Face Disguise Datasets}:} The training of the deep convolutional network used for facial key-point detection requires a large amount data. However, such datasets are not available (small: AR~\cite{ar} and Yale~\cite{yale} face databases) because of which researchers have relied upon transfer learning to detect facial key-points~\cite{stanford}. Transfer learning often performs well but may underperform if the data is not sufficient to fine-tune the pre-trained network. In order to avoid the above-mentioned issues, we proposed two simple and complex Face Disguise (FG) Datasets that can be used by researchers in the future to train deep networks for facial key-point detection. 

 \end{itemize}
 
 \begin{figure}[t!] 
\centering    
\includegraphics[scale = 0.23]{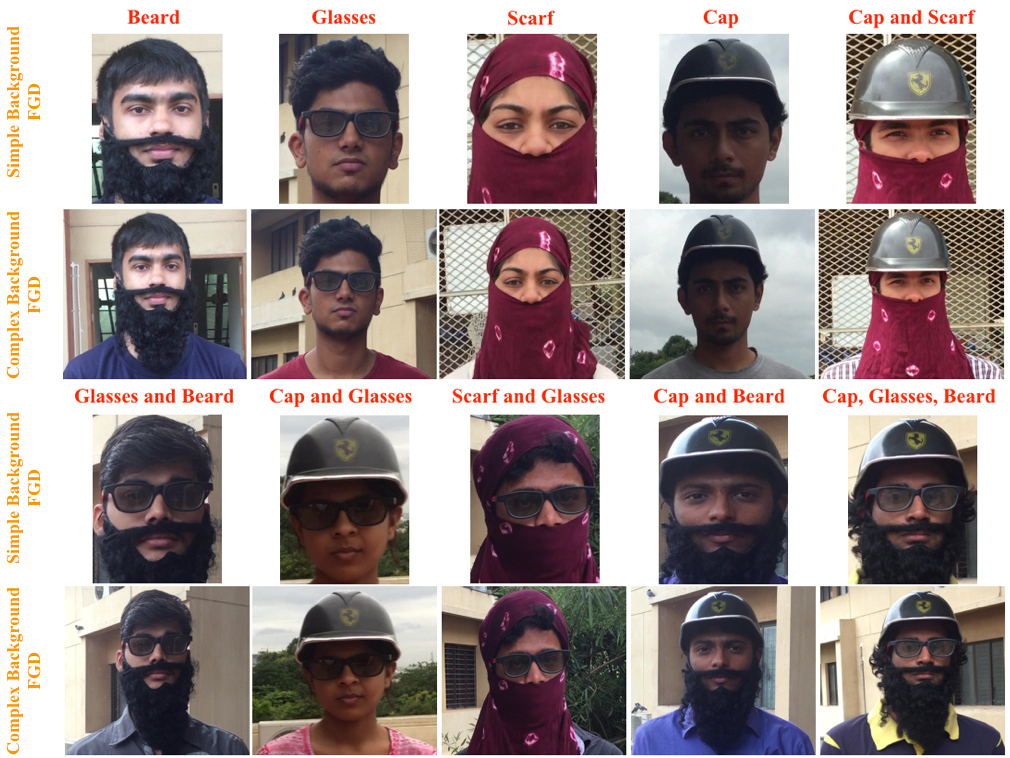}
\caption{{The illustration shows samples images with different disguises from both the Simple and Complex face disguise (FG) datasets. As seen from the image, the samples from the complex background dataset have a relatively complicated background as opposed to the simple dataset. }}
\label{fig:scatter00}
\end{figure}

\begin{figure*}[t!] 
\centering    
\includegraphics[scale = 0.49]{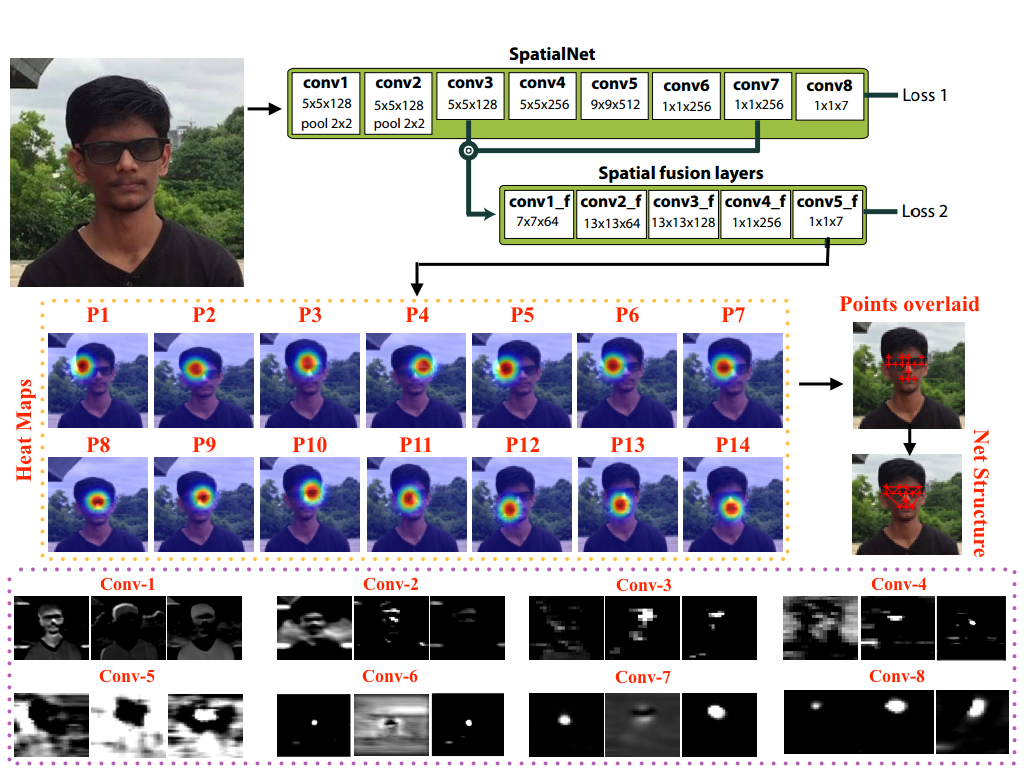}
\caption{{\textbf{Key-point detection pipeline}: The illustration shows the locations for all P1-P14 key-points generated as heat-maps by the spatial fusion convolution network. The figure also shows the net structure formed by combining the key-points. The above figure also shows the selected convolution layer activations for all 8 Conv layer of the network for the input image. }}
\label{fig:scatter00}
\end{figure*}

The proposed framework is used to perform facial disguise identification on the introduced datasets. The average key-point detection accuracy is presented for each key-point for both datasets. In addition, an extensive comparison of the proposed pipeline with other key-point detection methods is also presented. Finally, the effectiveness of the classification pipeline is also demonstrated with comparison with the state-of-the-art face disguise classification methods.

The paper is divided into the following sections. Section 2 presented the Datasets introduced in the paper while Section 3 presents the proposed Disguised Face Identification (DFI) Framework. Section 4 presents the experimental results and Section 5 draws conclusions.

\section{Simple and Complex Face Disguise Datasets}
The databases generally used for disguise related research (AR~\cite{ar} and Yale~\cite{yale} face databases) contain a small number of images with very limited disguise variations, such as scarves and/or sun-glasses. The deep learning networks to train require a large number of images with various combinations of disguises like people with glasses, beard, different hairstyles and scarf or cap. Therefore, we propose two face disguise (FG) datasets of 2000 images each with (i) Simple and (ii) Complex backgrounds that contain people with varied disguises, covering different backgrounds and under varied illuminations. Each proposed dataset (Simple and Complex) is formed of 2000 images recorded with male and female subjects aged from 18 years to 30 years. The dataset of disguised faces was collected in 8 different backgrounds, 25 subjects and 10 different disguises. The disguises in the dataset are namely:  (i) sun-glasses (ii) cap/hat (iii) scarf (iv) beard (v) glasses and cap (vi) glasses and scarf (vii) glasses and beard (viii) cap and  scarf (ix) cap and beard (x) cap, glasses, and scarf. The example images from each dataset are shown in Fig. 2

\section{Disguised Face Identification (DIC) Framework}
This section presents the introduced Disguised Face Identification (DIC) Framework. The DIC framework first detects the 14 facial key-points using the Spatial Fusion Convolutional Network~\cite{fusion}. Spatial Fusion Convolutional Network predicts and temporally aligns the facial key points of all neighbouring frames to a particular frame by warping backwards and forwards in time using tracks from dense optical flow. The confidence in the particular frame is strengthened with a set of 'expert opinions' ( with corresponding confidences)  from frames in the neighbourhood, from which the facial key points can be estimated accurately~\cite{fusion}. This makes the predictions from the Spatial Fusion Convolutional Network more accurate then other deep networks~\cite{fusion}. The detected points are connected to form a star-net structure as shown in Fig. 3. The detected points are next used by the proposed face identification approach (Section 3.2) to perform classification. 

\subsection{Facial KeyPoint Detection}
The key-point detection part of the DIC framework used the Spatial Fusion Convolutional Network~\cite{fusion} for key-point detection. The facial key-point detection problem is formulated as a regression problem that can be modeled by the Spatial Fusion Convolutional Network. The CNN takes an image and outputs the pixel coordinates of each key-point. The output of the last (conv-8) conv layer is a $i \times j \times k$ -dimensional cube (here 64 $\times$ 64 $\times$  14 k=14 key-points). 

The training objective is to estimate the network weights  $\lambda$ with the available training set $D=(x,y)$ and the regressor (conv8 output) is:
\begin{equation}
arg \min_{\lambda}\sum_{(x,y)\epsilon D}\sum_{i,j,k}||G_{i,j,k}(y_{k})-\phi_{i,j,k}(x,\lambda)||^{2}
\end{equation}

\begin{figure}[t!] 
\centering    
\includegraphics[scale = 0.33]{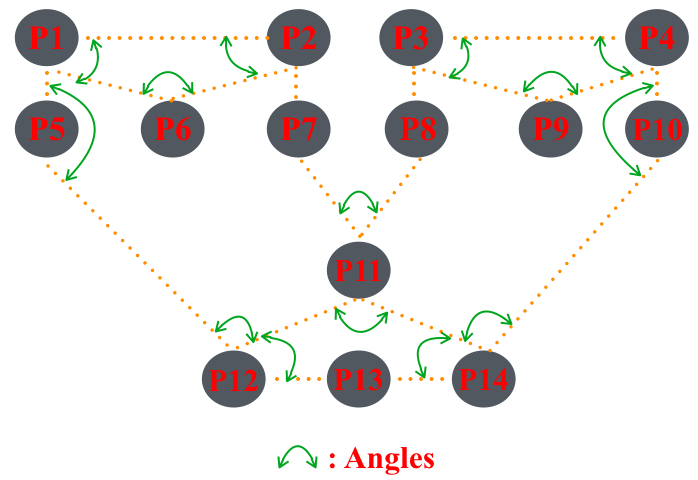}
\caption{{\textbf{Face Classification}: The figure shows the fashion in which the facial key-points are connected to form the net-star structure. The angles of this structure are later used to perform face identification.}}
\label{fig:scatter00}
\end{figure}

\begin{equation}
G_{i,j,k}(y_k)=\frac{1}{2\pi\sigma^{2}}e^{-[(y_k^1-i)^{2}+(y_k^2-j)^{2}]/2\sigma^{2}}
\end{equation}

is a Gaussian cantered at joint $y_k$.

    The ground truth labels are heat-maps synthesized for each key-point separately by placing a Gaussian with fixed variance at the ground truth key-point position $(x,y)$. The $l2$ loss penalizes the squared pixel-wise differences between the predicted heat-map and the ground truth heat-map. We used MatConvNet to train and validate the Fusion Convolutional Network~\cite{fusion} in MATLAB.

To create a net structure on the face, the locations produced by the above-explained network network for each point are connected as shown in Fig. 4.

\subsection{Disguised Face Classification}
In this section, we compare a disguised face to 5 non-disguised different person faces, including, the person in the disguised face. The classification is considered accurate if $\tau$ is minimum for the analysis between the disguised image and non-disguised image of the same person. 

The similarity of a disguised face is estimated against the non-disguised face by computing an L1 norm between the orientation of different key points obtained using the net-structure. In the net structure, the point at the nose is the reference point for the various angles that are to be measured as shown in Fig. 4. 

The similarity can be calculated according to the equation below:

\begin{equation}
\tau=\sum_i^ .\left|\theta_{i} - \phi_{i}\right|
\end{equation}

where $\tau$ is the similarity, $\theta_{i}$ represents the orientation of the $i^{th}$ key point of the disguised image, and $\phi_{i}$ stands for the corresponding angles in the non-disguised image. 

\begin{table}[!t]%
\centering
\caption{{Table shows the key-point detection accuracy (in \%) on the simple background and complex background, face disguise (FG) dataset. The accuracy is tabulated with respect to the distance $d$ (5, 10 and 15) in pixels from the ground truth (GT). There are 14 rows corresponding to 14 facial key points and the last row corresponds to average of all the facial key points plots.}}
            \begin{tabular}{>{}m{1cm}|c c c|cccc}
\hline
\multicolumn{1}{c}{Points} & \multicolumn{6}{c}{Distance (Pixels) from Ground Truth (GT)}   \\ \hline
\multicolumn{1}{c}{--} & \multicolumn{3}{c}{Simple (FG) Dataset} & \multicolumn{4}{c}{Complex (FG) Dataset} \\ \hline
    & d = 5 & d = 10 & d = 15 & d = 5 & d = 10 & d = 15 \\
\cline{2-4} \hline
P1 & 54 & 86  & 97 & 32 & 68 & 90 \\ 
P2 &  85 & 95 & 98 & 84 & 94 & 97 \\
P3 & 85 & 100 & 100 & 74 & 97 & 97 \\
P4 & 83 & 99 & 100 & 64 & 93 & 94 \\
P5 & 82 & 96 & 96 & 64 & 90 & 94  \\
P6 & 87 & 98 & 99 & 85 & 98 & 99 \\
P7 & 40 & 78 & 97 & 36 & 75 & 96 \\
P8 & 82 & 99 & 99 & 74 & 99 & 99 \\
P9 & 39 & 75 & 95 & 32 & 70 & 95 \\
P10 & 93 & 97 & 97 & 64 & 96 & 96 \\
P11 & 97 & 99 & 99 & 96 & 99 & 99 \\
P12 & 54 & 84 & 94 & 41 & 74 & 90 \\
P13 & 91 & 96 & 96 & 85 & 93 & 93 \\
P14 & 73 & 95 & 95 & 46 & 76 & 89 \\
\hline
\hline
All & 85 & 94 & 94 & 56 & 89 & 92 \\
\hline
\end{tabular}
\end{table}


\begin{figure*}[t!] 
\centering    
\includegraphics[scale = 0.49]{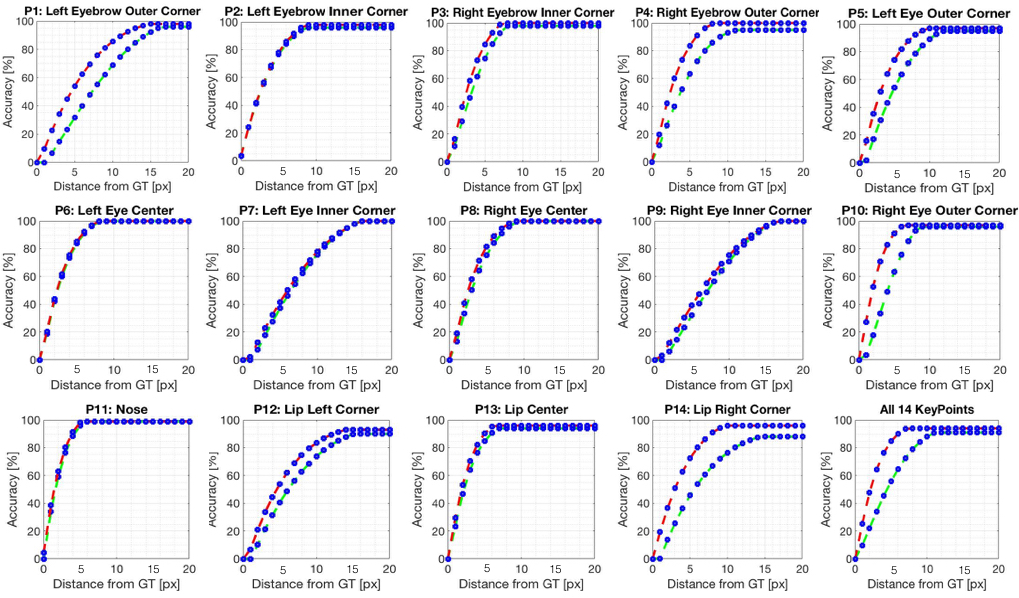}
\caption{{The figure shows the key-point detection performance graphs for all P1-P14 facial key-points for both simple and complex face disguise (FG) datasets. The green color represents the key-point accuracy computed on the complex dataset while red color represents the accuracy on the simple dataset.}}
\label{fig:scatter00}
\end{figure*}

\section{Experimental Results}
This section presents the experiments performed on the two introduced datasets namely: (i) Simple background face disguise (FG) dataset (ii) Complex background face disguise (FG) dataset, using the disguised face identification (DFI) framework. The training of the key-point detection spatial fusion network module of the DFI framework is next described followed by the evaluation protocol used to evaluate the effectiveness of the network. The key-point detection performance of each key-point along with a comparison with other deep learning architectures is presented. The classification performance of the DFI framework is also presented for different face disguises along with a comparison with the state of the art. 

\subsection{Spatial Fusion ConvNet Training}
We train the Spatial Fusion CNN on 1000 training images, 500 validation images and 500 test images, randomly selected from the disguised faces dataset (simple and complex, trained separately). We trained the network for 90 iterations using a batch size of 20. A 248$\times$248 sub-image is randomly cropped from every input image. the cropped patch is randomly flipped, randomly rotated between $-40^{\circ}$ and $+40^{\circ}$, and resized to 256$\times$256 before given as input to the network for training. The variance of the Gaussian is set to 1.5 with an output heat-map size of 64$\times$64 (Section 3.1). The base learning rate is $10^{-5}$, which we decrease to $10^{-6}$ after 20 iterations. Momentum is set to 0.9. 

\subsection{Key-Point Detection Performance}
In this section, we analyze the performance of key point detection spatial fusion network module of the proposed disguised face identification (DFI) framework on the both datasets introduced in this paper. The performance of the module is evaluated by comparing the coordinates of the detected key-points with their ground truth values in the annotated dataset. 

We have presented the performance of the key-point detection spatial fusion network in the form of graphs that plot accuracy vs distance from the ground truth pixels, where a key point is deemed correctly located if it is within a set distance of $d$ pixels from a marked key point center in ground truth. The key-point detection performance for both the simple (red) and complex (green) background face disguise dataset is plotted for each key-point as shown in Fig. 5. As we can see the accuracy increases as the distance from the ground truth pixel increases. 

Table 1 provides the quantitative comparison of the predicted key-points for both the datasets at 3 (d = 5, 10, 15) pixel distances from the ground-truth. As observed for d = 5, an average key-point detection accuracy of 85\% was recorded for the simple background dataset as opposed to an accuracy of 74\% for the complex background dataset. The accuracy increases for both datasets with an increase in pixel distance from the ground-truth.    

\begin{figure}[t!] 
\centering    
\includegraphics[width = \linewidth, height = 6.5 cm]{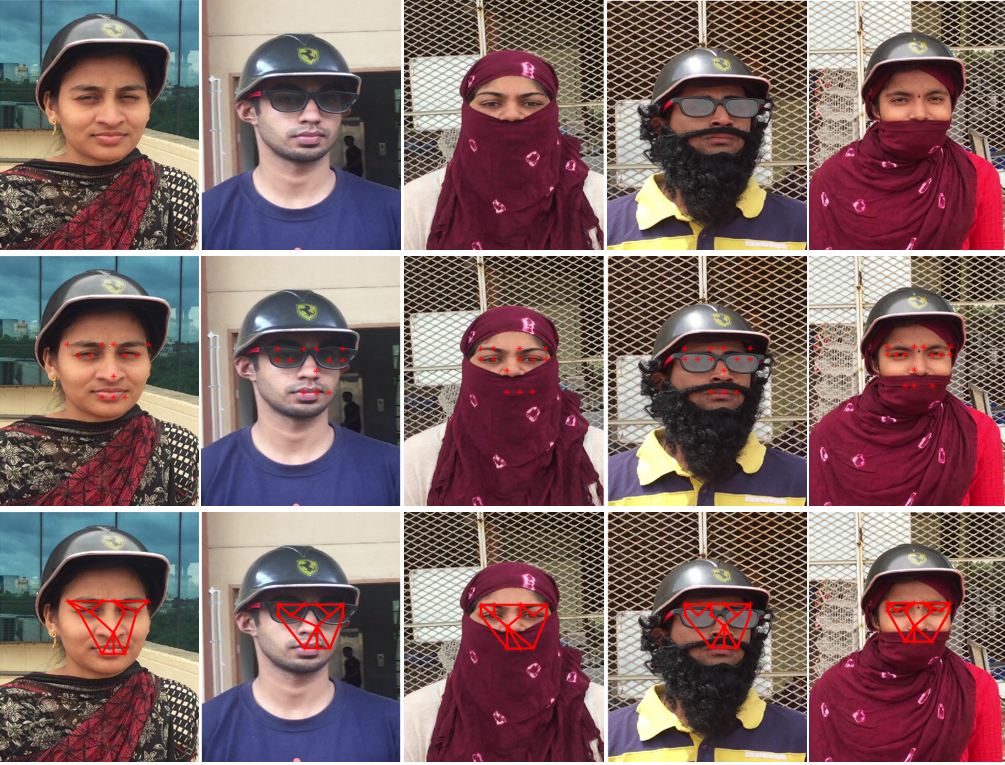}
\caption{{The figure shows the key-points and the net-structure captured using the spatial fusion network in the DFI framework on the image samples chosen from the complex background dataset. The first row consists of 5 disguises which are given as inputs to the network, the second row shows the detected facial key-points while the third row consists of the representation of the net structure formed by joining different key points. }}
\label{fig:scatter00}
\end{figure}

The lower performance on the complex dataset is observed as the proposed network occasionally fails to detection key-points that are in the outer region of the face as they seem to be affected by the preference of background clutter as shown in Fig. 6. Upon close observation, it was found that the outer facial key points – P1, P4, P5, P10, P12, P14 showed a similar or lower accuracy than the inner key points - P2, P3, P6, P7, P8, P9, P11 for the same pixel distance for the complex dataset. For example, the point-P1 showed a lower accuracy of 32\%, 68\% and 90\% on the complex background dataset as compared to the 96\%, 99\% and 99\% accuracy on the simple background dataset as recorded for 5, 10 and 15 pixels distance from the ground-truth, respectively. The above stated is also visually supported in Fig. 7.

\subsection{Key-Point Performance Analysis wrt. Background Clutter }
This section further analyzes the effect of background clutter as it significantly affects the key-point detection performance as observed in the previous section. The effect of background clutter is observed by analyzing the key-point detection performance of the points in the eye, nose and lip regions. 

\begin{figure}[t!] 
\centering    
\includegraphics[scale = 0.27]{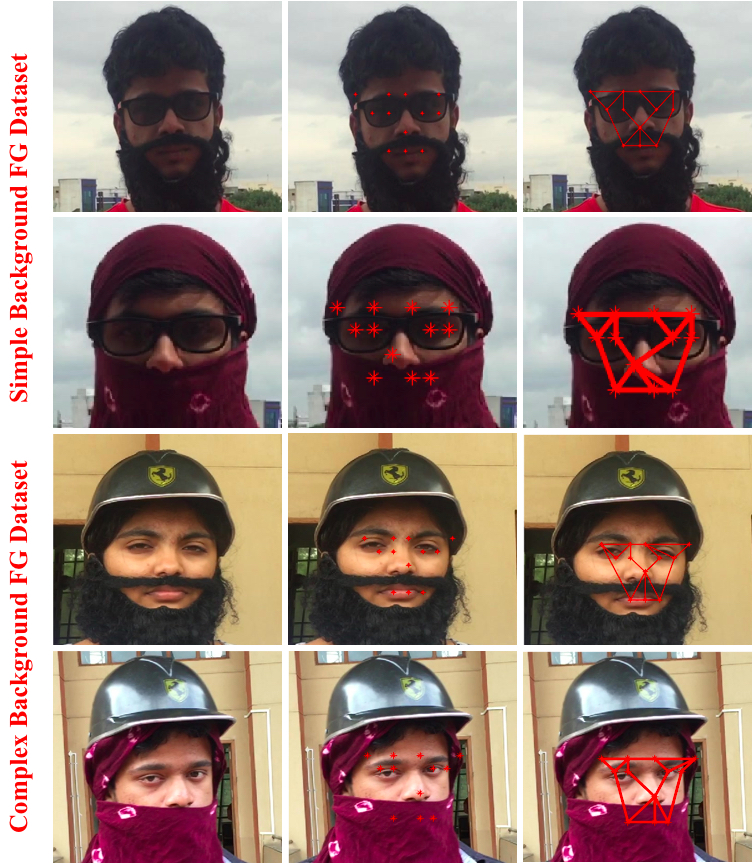}
\caption{{\textbf{Outer Key-Point Analysis}: Figure shows two sample images from the simple and complex FG dataset along with the detected key-points as well as the net-star structure. It is observed from the examples that the DIC framework captures outer region key-points (P1, P4, P5, P10, P11, P13) more accurately in the simple dataset as opposed to the complex dataset. }}
\label{fig:scatter00}
\end{figure}

\subsubsection{Eye Region Key-Points Detection Performance}
This section analyzes the key-point detection performance of the spatial fusion network in the DFI framework for the key-points in the eye region for both simple and complex background datasets. The eye facial key points include the points P1, P2, P3, P4, P5, P6, P7, P8, P9, and P10. As observed from Fig. 5 at a particular distance, the performance of the key-point detection performance is better for the simple dataset as opposed to the complex background dataset. The effect of background clutter is prominent for the points on the outer region of the eyes namely: P1, P4, P5, and P10. In addition, the accuracy at pixel distance closer to the ground-truth is significantly higher for the simple dataset further emphasizing the damaging effect of background clutter on key-point detection accuracy.

\subsubsection{Nose Key-Point Detection Performance}
The nose key-point (P11) is detected with similar accuracy for both plane and complex background dataset as shown in Fig. 5. This point is placed in the center of the face and doesn't seem to be affected by the background clutter. 

\subsubsection{Lips Region Key-Points Detection Performance}
The lip region consists of P12, P13, and P14 key-points. The detection accuracy of P12 and P14 is affected by the presence of background clutter as seen from the large difference between the red and green line as observed at a particular distance from Fig. 5. However, the key-point detection performance of P13 seems to be similar for both the simple and complex background datasets. The reason for the poor key-point detection performance of P12 and P14 seems to be due to their outer location. 

\subsection{Facial Key-points Detection: Multiple Persons}
In this section we analyze the performance of our model on the images containing multiple faces. As our model is trained on images containing a single person with cluttered background, we use a viola jones face detector to first locate multiple faces from the given image. The proposed DIC framework is used on each face to extract the net structure for each individual face as shown in Fig. 8. The key-point detection classification performance for each simple and complex datasets for 2 faces in the image are 80\% and 50\% while for 3 faces in the image are 76\% and 43\% respectively. There is a decrease in accuracy as compared to the 85\% and 56\% key-point detection performance of simple and complex background dataset for a single face in the image.

\begin{figure}[!t]
\begin{subfigure}{0.5\textwidth}
  \centering
   \includegraphics[scale = 0.27]{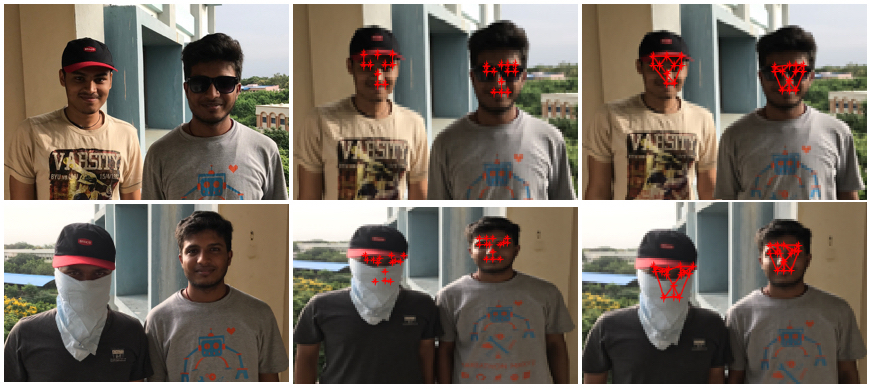}
  \label{fig:sfig1}
\end{subfigure}
\begin{subfigure}{0.48\textwidth}
  \centering
 \includegraphics[width = \linewidth, height = 3.3 cm]{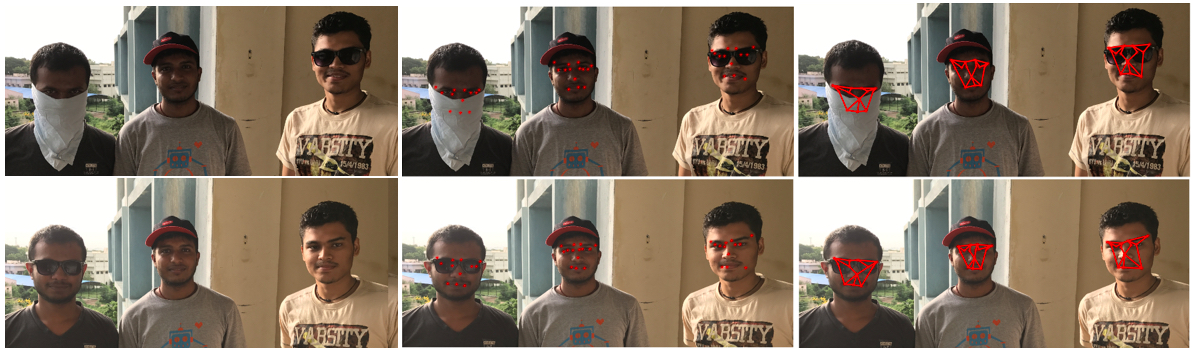}
  \label{fig:sfig2}
\end{subfigure}

\caption{{Figure shows the key-points and the net-structure captured using the spatial fusion network module in the DIC framework for more than one person within the image.}}
\label{fig:fig}
\end{figure}

\subsection{Comparison of KeyPoint Detection Performance with Other Architecture}
In this section we compare the key-point detection performances of the Spatial Fusion Convolutional Network~\cite{fusion} used in this paper with various other architectures as shown in Table 2. We have considered 3 architectures namely CoordinateNet (CN), CoordinateNet extended(CNE) and SpatialNet~\cite{fusion}. The key-point detection accuracies for the above-mentioned networks for the above-mentioned architectures is presented for the simple background face disguise dataset and complex face disguise dataset at d = 5 pixel distance from the ground-truth. The key-point detection accuracy results for simple background are 77.6\%, 78.2\%, 81\% and 85\% for CN, CNE, SpatialNet and Spatial Fusion~\cite{fusion} network in the proposed DIC framework, respectively. The Spatial Fusion network outperforms the other networks by a significant margin. The classification results for complex background face disguise dataset are 44\%, 44.7\%, 52.67\% and 56\% for SpatialNet and Spatial Fusion network, respectively. We can clearly see that when the background is cluttered the accuracy decreases drastically.

\begin{table}[!h]%
\centering
\caption{{Comparison of classification accuracies(\%) of various architectures namely Coordinate Net (CN) \cite{fusion}, Coordinate extended (CNE) \cite{fusion}, Spatial net and Spatial Fusion~\cite{fusion} (DFI) on the simple and complex face disguise datasets.}}
            \begin{tabular}{>{}m{1.60cm}|ccccc}
\hline
\multicolumn{1}{c}{Dataset} & \multicolumn{4}{c}{Other Architectures}   \\ 
\hline
    &  DFI & CN  & CNE & Spatial Net   \\
\cline{2-4} \hline
\small{Simple} & \cellcolor{gray!50}85& 77.6 & 78.2 & 81 \\ 
\hline
\small{Complex} & \cellcolor{gray!50}56& 44 & 44.7 & 52.67 \\ 
\hline

\end{tabular}
\end{table}

\subsection{Classification Performance and comparison with the state-of-the-art}
This section presents the disguise face classification performance for each disguise for both the simple and complex datasets. It is observed from Table. 3 that the facial disguise classification performance decreases with an increase in the complexity of the disguise. 

\begin{table}[!h]%
\centering
 \caption{Table presents the face disguise classification accuracies(\%) for selected disguises on both datasets. }
            \begin{tabular}{>{}m{0.25cm}|ccccc}
\hline
\multicolumn{1}{c}{Dataset} & \multicolumn{4}{c}{Disguises}   \\ 
\hline
    &  {cap}  & {scarf} & {cap + scarf} & {cap + glasses + scarf} \\
\cline{2-4} \hline
\small{Simple} &  90 & 77 & 69 & 55 \\ 
\hline
\small{Complex} & 83 & 67 & 56 & 43 \\ 
\hline

\end{tabular}
\end{table}

The disguise face performance is compared with the state-of-the-art disguise face identification approach~\cite{tejas}. The proposed classification framework was able to outperform the \cite{tejas} by 13\% and 9\% respectively, as shown in Table. 4.

\begin{table}[!h]%
\centering
\caption{Table shows the face disguise classification accuracy (\%) compared against the state-of-the-art \cite{tejas}}
            \begin{tabular}{>{}m{1.60cm}|c| c }
\hline
\multicolumn{1}{c}{Dataset} & \multicolumn{2}{c}{Comparison}   \\ 
\hline
    & DFI & state-of-the-art \cite{tejas}  \\
\cline{2-3} \hline
\small{Simple FG Dataset} & \cellcolor{gray!50}\textbf{78.4} & 65.2 \\ \hline
\small{Complex FG Dataset} & \cellcolor{gray!50}\textbf{62.6} & 53.4 \\ 
\hline
\end{tabular}
\end{table}

\section{Conclusion}
The paper presents the Disguised face identification (DFI) framework that first detects the facial key-points and then uses them to perform face identification. The framework is evaluated on two facial disguise (FG) datasets with simple and complex, introduced in the paper. The framework is shown to outperform the state-of-the-art methods on key-point detection and face disguise classification. The large number of images and disguised in the introduced datasets will improve the training of deep learning networks avoiding the need to perform transfer learning.  

\bibliographystyle{ieee}
\bibliography{egbib}

\end{document}